\documentclass{article}

\usepackage{arxiv}
\usepackage{graphicx, amsmath, pgfplots, amsfonts, setspace, hyperref, natbib}
\usepackage{algorithm}
\usepackage{algpseudocode}
\DeclareMathOperator*{\argmax}{argmax}
\renewcommand{\vec}[1]{\mathbf{#1}}
\pgfplotsset{compat=1.17}

\title{Bayes Classification using an approximation to the Joint Probability Distribution of the Attributes}

\author{ \href{https://orcid.org/0000-0003-1729-559X}{\includegraphics[scale=0.06]{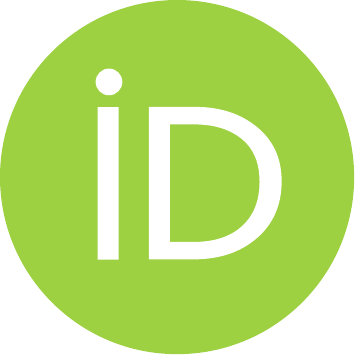}\hspace{1mm}Patrick Hosein}\\
	Department of Computer Science\\
	The University of the West Indies\\
	St. Augustine, Trinidad\\
	\texttt{patrick.hosein@sta.uwi.edu} \\
	\And
	\href{https://orcid.org/0000-0003-0612-7726}{\includegraphics[scale=0.06]{orcid.pdf}\hspace{1mm}Kevin Baboolal} \\
	Department of Computer Science\\
	The University of the West Indies\\
	St. Augustine, Trinidad\\
	\texttt{k.baboolal1.0@gmail.com} \\
}

\renewcommand{\shorttitle}{Bayes Classifier}

\hypersetup{
pdftitle={Bayes Classification using Conditional Probability approximations},
pdfauthor={Patrick Hosein and Kevin Baboolal},
colorlinks=true
}

\begin{document}
\maketitle   

\begin{abstract}
The Naive-Bayes classifier is widely used due to its simplicity, speed and accuracy. However this approach fails when, for at least one attribute value in a test sample, there are no corresponding training samples with that attribute value. This is known as the zero frequency problem and is typically addressed using Laplace Smoothing. However, Laplace Smoothing does not take into account the statistical characteristics of the neighbourhood of the attribute values of the test sample. Gaussian Naive Bayes addresses this but the resulting Gaussian model is formed from global information. We instead propose an approach that estimates conditional probabilities using information in the neighbourhood of the test sample. In this case we no longer need to make the assumption of independence of attribute values and hence consider the joint probability distribution conditioned on the given class which means our approach (unlike the Gaussian and Laplace approaches) takes into consideration dependencies among the attribute values. We illustrate the performance of the proposed approach on a wide range of datasets taken from the University of California at Irvine (UCI) Machine Learning Repository. We also include results for the $k$-NN classifier and demonstrate that the proposed approach is simple, robust and outperforms standard approaches.


\end{abstract}

\section{Introduction}

Machine Learning Classifiers are regularly used to solve a wide range of classification problems. Several techniques are available ($k$-NN, XG-Boost, Singular Value Decomposition, Neural Networks, etc.) but we focus on the Naive Bayes Classifier. This classifier is simple, the results can be interpreted (i.e., explainable) and its performance is quite good. However, it has two major drawbacks, the assumption of independence of attribute values and the zero-frequency problem. The latter problem is normally addressed using Laplace Smoothing or through the use of Gaussian Naive Bayes. We propose an approach in which the zero frequency problem is avoided by instead approximating the conditional probability using samples in the neighbourhood of the test sample. In fact our approach does not need to make the independence assumption used for Naive Bayes and so is able to capture dependencies among attributes.

In the next section we describe related work and compare those with our contributions. We also illustrate our classifier using a simple illustrative example. We then use several datasets to compare the proposed classifier with Naive Bayes with Laplace Smoothing, Gaussian Naive Bayes and finally the $k$-NN Classifier. Finally we discuss issues such as computational complexities and why we believe that our classifier may be of benefit to the community.

\section{Related Work and Contributions}

We first note that our approach is related to a Parzen-Window estimator (\cite{parzen1962estimation}) but with two differences. Firstly, in our approach there is no dependence on the number of samples (as with the Parzen-Window Kernel). Secondly, our approach uses a single hyper-parameter unlike the non-parametric Parzen-Window estimator. However, we will show that the optimal parameter value of our approach is dependent on the number of samples (as well as other properties of the dataset) and so in future work we will investigate this dependency. The corresponding Parzen-Window Kernel in our approach, for given test sample $\vec{x}$ and training sample $\vec{x}_i$ with $n$ samples can be written in this form
\begin{equation}
\delta_n = \frac{1}{(||\vec{x} - \vec{x}_i||_2)^{\kappa}}
\end{equation}
where $\kappa>0$ is the tuning parameter.
The results in this paper are obtained by searching over $\kappa$ since we are still in the process of determining its dependence on $n$. 

The related literature on Naive-Bayes classification can generally be sorted into four categories - structure extension, feature selection, attribute weighting and  local learning. As described previously, one drawback of the Naive-Bayes algorithm is the attribute independence assumption. Structure extension tackles this problem by determining feature or attribute dependencies. Often, approaches utilize the factor graph representation of probability distributions to depict the Bayesian network. Learning the optimal Bayesian network is regarded as an intractable problem and researchers usually employ techniques to reduce the complexity. One such method is described in \cite{article} where the author uses a directed graph to represent the distribution. They refer to this method as Tree Augmented Naive Bayes or TAN. The structure of the TAN graph is limited in that features can have at most two parents, one parent must be the class and the other parent can be an attribute. These constraints reduce the complexity of learning and the problem formulates as $p(C=c|\vec{X}=[X_1,X_2, \ldots, X_n])$ with learning occurring in polynomial time. TAN allows for edges between child vertices (attributes), and hence creates dependencies between attributes, reducing the independence assumption of the Naive-Bayes classifier.  

An extension of the TAN method is developed in \cite{4721435} and is called the Hidden Naive Bayes method. The authors add a parent vertex to every attribute called a hidden parent. The degree of the vertex is maintained at two thus limiting computational complexity. The hidden parent is computed such that it represents the influences of all other attributes and a weight is used to represent the importance of an attribute. The optimal class is the one which maximizes $p(c)*\Pi p(a_i|a_{hpi},c)$ where $a_{hpi}$ represents the calculated hidden parent. $\Pi \ p(a_i|a_{hpi},c)$ is given by $\Pi \  p(a_i|a_j,c)*w_{i,j} $. The weights $w_{i,j} $ are calculated from the conditional mutual information of $p(a_i;a_j|C)$. 

The authors in \cite{math9222982} builds on the work of the Hidden Naive Bayes model and uses instance weighting to achieve higher accuracy. The Hidden Naive Bayes model treats every instance with equal importance which may not always hold true as different instances could have different contributions. They create discriminative instance weights to improve performance. Weights are applied in the calculation of the conditional mutual information of $p(X_i;X_j|C)$ and the prior probabilities $p(C)$. The authors opt for an eager learning method to reduce computational complexity and use frequency-based instance weighting.

Feature selection is characterized by the determination of a feature subset based on the available feature space. Feature selection is important for the removal of redundant or uninformative features in the Naive-Bayes algorithm, which is particularly sensitive to redundant features by virtue of its computation. To demonstrate this, take a Naive Bayes classifier with two features $[X_1, X_2]$. The Classifier function to be maximized can be written as $p(C_i)p(X_1|C_i)p(X_2|C_i)$ if we add another feature that is redundant to feature $X_2$ our classifier function is now given by $p(C_i)p(X_1|C_i)p(X_2|C_i)^2$, therefore the classifier becomes biased.  The class of Naive-Bayes algorithms that adopt feature selection for performance boost is mainly a Naive-Bayes algorithm with a prepossessing step that improves the classifiers performance. \cite{Langley1994InductionOS} explores the method of feature selection, with a greedy search algorithm through the feature space that seeks to eliminate redundant attributes and improve the overall accuracy.    

Attribute weighting involves the assignment of weights to features to mitigate the independence assumption of the Naive-Bayes classifier. This can be accomplished through scoring of the attributes with a classifier for evaluation. Alternatively, heuristics can be used to determine the characteristics of the attributes and an associated weight. In \cite{6137329} the authors present one such application where the classification decision is given from maximizing  $p(c)\Pi p(a_i|c)^{w_i}$. The attribute weight, $w_i$ is determined from the Kullback-Leibler divergence and represents the dissimilarity between the a-priori and  posteriori probabilities. A high dissimilarity means that a feature would be useful. Building on the work of \cite{6137329} the authors of  \cite{cmc.2022.022011} use  an exponential weighting scheme per feature. 

Local learning employs the use of localized data to compute probabilities of the Naive-Bayes algorithm. \cite{10.1007/3-540-47887-6_10} computes multiple Naive-Bayes classifiers for multiple neighborhoods using different radii from the test object. The most accurate Naive-Bayes classifier is then used for classifying the new object. The aforementioned method bears close similarity to the $K$-Nearest Neighbors ($k$-NN) classification but differs from $k$-NN by replacing the majority vote classification with a Naive-Bayes classifier. The method presented in our paper also makes use of distance similarity as in $k$-NN but our method remains generative whereas $k$-NN is discriminative. In \cite{Gweon2019TheKC} the authors also use distance information to estimate conditional probabilities but they only consider the $k$ nearest neighbours as in the $k$-NN classifier and no weighting is performed as in our approach. They show accuracy improvements for some datasets when compared to other models such as $k$-NN. 

Another example of locally weighted learning is presented in \cite{10.5555/2100584.2100614} in which they use a nearest neighborhood algorithm to weight the data points. The neighborhood size is defined by the user but results were found to be relatively insensitive to this choice. Data points within the neighborhood are linearly weighted based on the euclidean distance. The weights for each data point are then used when computing the prior probability $p(C=c)$ and the conditional probabilities $p(\vec{X}=\vec{x}|C=c)$. Data should be normalized for use in this method. 

\section{Standard Naive-Bayes Classifiers}

We illustrate how the Naive-Bayes classifier uses training data to estimate conditional probabilities and these can then be used to classify unknown samples (the test set) \cite{10.1007/978-3-540-77046-6_2}. Let $C$ denote the random variable representing the class of an instance and let $\vec{X} = [X_1, X_2, \ldots, X_K]$ be a vector of random variables denoting the set of possible attribute values. Let $c$ represent a particular class belonging to $\vec{c}$ and let $\vec{x} = [x_1, x_2, \ldots, x_K]$ represent the attribute values of a specific sample. For a given test sample $\vec{x}$ we can use Bayes Theorem to obtain 
\begin{equation}
    p(C = c | \vec{X} = \vec{x}) = \frac{p(C = c) p(\vec{X} = \vec{x} | C = c)}{p(\vec{X} = \vec{x})}
\end{equation}
 If we assume that the attribute values are independent then we have
\begin{equation}
     p(\vec{X} = \vec{x} | C = c) = \prod_{i=1}^K p(X_i = x_i|C =c)
\end{equation}
Finally using the fact that the term $p(\vec{X}=\vec{x})$ is invariant across the classes then one can find the most probable class $c^*$ (and hence the predicted class) as
\begin{equation}
c^* = \argmax_{c \in \vec{c}} \left\{  p(C = c) \prod_{i=1}^K p(X_i = x_i|C =c) \right\}
\end{equation}
The conditional probabilities $p(X_i=x_i|C = c)$ and the probability $p(C = c)$ are estimated from the training data. 

\subsection{Laplace Smoothing}

The conditional probability $p(X_i = x_i|C = c)$ is computed as the ratio of the number of instances (in the training data) in which that attribute value $x_i$ occurs and the sample is tagged as $c$ and the total number of instances of $c$. 
This is the typical approach for estimating this probability. However if this probability is zero for at least one attribute $i$ then the product over all attributes will be zero. To avoid this from occurring the Laplace-Estimate and the $M$-Estimate is used. Using these estimates we instead have
\begin{equation}
p(C = c) = \frac{n_c + k}{N + k|\vec{c}|}    
\end{equation}
where $n_c$ is the number of instances satisfying $C = c$, $N$ is the number of training instances, $|\vec{c}|$ is the number of classes and typically $k=1$ is used. In $M$-Estimate, 
\begin{equation}
p(X_i = x_i|C = c) = \frac{n_{ci} + mq}{n_c + m}
\end{equation}
where $n_{ci}$ is the number of instances where $X_i = x_i$ and $C = c$, $n_c$ is the number of instances satisfying $C = c$, $q = p(X_i = x_i)$ is the prior probability of $x_i$ estimated by the Laplace-Estimate and typically $m=1$.

\subsection{Gaussian Naive Bayes}

In the case of Cardinal attributes $p(X_i = x_i|C = c)$ is typically modelled by some continuous probability distribution (e.g., Gaussian). In this case
\begin{equation}
p(X_i = x_i|C = c) = \frac{1}{\sqrt{2 \pi \sigma^2}} e^{\frac{-(x_i - \mu)^2}{2 \sigma^2}}
\end{equation}
where
\begin{equation}
    \mu = \frac{1}{n_c} \sum_{\{i|C = c\}} x_i \qquad{\sf and}\qquad \sigma^2 = \frac{1}{n_c} \sum_{\{i|C = c\}} (x_i - \mu)^2
\end{equation}

\subsection{Discretization}

Another approach to address the zero occurrence issue above is to cluster consecutive attribute values and instead use the cluster in the analysis. With sufficiently large cluster sizes the probability that an attribute value does not belong to a cluster is significantly reduced. Equal width discretization is a typical approach used. For each attribute, $X_i$, a distinct (non-overlapping) interval $(a_i, b_i]$ is defined such that the conditional probabilities become
\begin{equation}
    p(X_i=x_i|C=c_i) \approx p(a_i < x_i \leq b_i|C = c)
\end{equation}
With sufficiently large intervals the probability that an attribute value for a given class has no samples in an interval is significantly small.

\section{Proposed Bayes Classifier}

Naive Bayes classifiers make the assumption that features are independent. Note that this assumption is made to reduce the zero-frequency problem. We do not make this assumption but instead approximate the conditional probability of the sample attributes set given a class. We first encode any categorical data into integer values so that we can compute a distance between the attribute values of any pair of samples. However, when computing such distances we normalize by the maximum range of differences for the given training set. Let $X_{ij}$ denote the value of attribute $j$ of training sample $i$. We define the distance $d_{ab}$ between two samples $a$ and $b$ as
\begin{equation}
    d_{ab} \equiv \left( \sum_{j=1}^n \left( \frac{X_{aj} - X_{bj}}{r_j} \right) ^2 \right)^{\frac{1}{2}}
\end{equation}
where $n$ is the number of attributes and $r_j$ is the range of values for attribute $j$ given by
\begin{equation}
    r_j \equiv \max_i \{X_{ij}\} - \min_i \{X_{ij}\}
\end{equation}

We approximate the conditional probability by using information in the neighbourhood of an unknown sample. Consider a test sample $\vec{x}$ with value $x_j$ for attribute $j$ and denote the distance between this sample and training sample $i$ by $d_i$. Let $y_i$ denote the class of training sample $i$. For a given class $c$ the probability that $\vec{x}$ occurs given $c$ is approximated by
\begin{equation}
    p(\vec{X}=\vec{x}|C=c) \approx \frac{1}{N_c} \sum_{\{i|y_i=c\}} (1 + d_i)^{-\kappa}
\end{equation}
where we use $N_c$ to denote the number of training samples of class $c$. The parameter $\kappa \geq 0$ can be tuned based on the dataset but we will soon see that, even if we use a constant $\kappa$ for all datasets, the performance is robust.

If $\kappa=0$ then the summation is equal to $N_c$ and so the approximation is equal to 1 since we will simply be summing all samples that belong to class $c$ which is given by $N_c$. If $\kappa$ is very large then all terms in the summation will be close to zero except for the samples that have the exact attribute values as the test sample. In this case the result is the relative frequency of the test sample in the training set. In most cases there will be no training samples identical to the test sample and so the result will be zero (i.e., the zero frequency problem). By proper choice of $\kappa$ we can capture a sufficient number of ``neighbourhood" samples to estimate the conditional probability. 

Note that, as the number of samples increases to infinity, $p(\vec{X}=\vec{x}|C=c)$ will approach the conditional distribution function and so when evaluated at the specified point we get $f(\vec{x}|c)$. Furthermore, as $\kappa$ goes to infinity $p(\vec{X}=\vec{x}|C=c)$ converges to the number of samples with attributes $\vec{x}$ over the total number of samples and hence converges to $f(\vec{x}|c)$. Therefore the proposed classifier approaches the optimal Bayes Classifier as the number of samples goes to infinity and $\kappa$ is taken to infinity.

We need to multiply by $p(c)$ before finding the predicted class. We therefore have
\begin{equation}
p(c) p(\vec{X}=\vec{x}|C=c) \approx \frac{N_c}{m} \frac{1}{N_c} \sum_{\{i|y_i=c\}} (1 + d_i)^{-\kappa}
\end{equation}
where $m$ is the number of samples. Since $m$ is constant then when taking the maximization over $c$ we can ignore the term $1/m$.
The pseudo-code for the proposed classifier is provided in Algorithm 1.

\begin{algorithm}
\caption{Pseudo-code for proposed Algorithm}\label{alg}
\setstretch{2}
\begin{algorithmic}
\State $\vec{c} \equiv \text{\sf set of classes}$
\State $\kappa > 0$ \text{tuning parameter}
\State $\vec{X} \in \mathbb{R}^{m \times n}$ \text{\sf ($m$ training samples with $n$ attributes)}
\State $\vec{y} \in \vec{c}^{m \times 1}$ \text{\sf (class of training samples)}
\State $\vec{x} \in \mathbb{R}^{1 \times n}$ \text{\sf (test sample attributes)}
\State $\displaystyle r_j \equiv \max_{i} \{X_{ij}\} - \min_i\{X_{ij}\} \qquad j = 1,\ldots,n$
\State $\displaystyle d_i \equiv \left( \sum_{j=1}^n \left( \frac{X_{ij} - x_j}{r_j} \right) ^2 \right)^{\frac{1}{2}}$
\State $\displaystyle c^*(\kappa) = \argmax_{c\in\vec{c}} \left\{ \sum_{\{i | y_i=c\}} (1 + d_i)^{-\kappa} \right\}$
\end{algorithmic}
\end{algorithm}

\section{Illustrative Example}
 
 Let us illustrate with a simple example of two classes (red and green) and a single attribute. There are four samples tagged as green with attribute values $X = 4, 5, 10, 11$. There are also four samples tagged as red with attribute values $X = 1, 2, 7, 8$.
 Let us suppose that all samples except the sample at $X=4$ are training samples and we will use the sample at $X=4$ for testing. 
 If we try regular Naive Bayes then $p(C=green|X=4) = 0$ because $p(X=4|C=green)=0$. Similarly $p(C=red|X=4)=0$. Hence it is not possible to choose a class.
 
 Next consider Naive Bayes with Laplace Smoothing. We have 
\[
 p(C = green) = (3 + 1)/(7 + 2) = 4/9 
 \qquad {\sf and} \qquad
 p(C = red) = (4 + 1)/(7 + 2) = 5/9
\]
Therefore we have,
\[
p(X = 4|C = green) = (0 + 1)/(3 + 7) = 1/10 
\]
\[
p(X = 4|C = red) = (0 + 1)/(4 + 7) = 1/11
\]
Therefore $p(\vec{X}=x) p(C= green|X = 4) = 4/90$ and $p(\vec{X}=x)(p(C = red | X=4) = 5/99$ and so we would choose RED which is incorrect.

Let us consider Gaussian Naive Bayes. In this case the green and red class samples are modelled by Gaussian distributions with 
\begin{equation}
    \mu_{g} = 8.67 \qquad {\sf and} \qquad \sigma_g = 2.6247
\end{equation}
and
\[
\mu_r = 4.5 \qquad {\sf and} \qquad \sigma_r = 3.041
\]
Therefore 
\[
p(C = green | X = 4) = 0.0315 \times 4/9 = 0.014
\]
and 
\[
p(C = red | X = 4)= 0.1293 \times 5/9 = 0.0718
\]
and hence class red is incorrectly chosen. Let us now consider the proposed approach.
We have
\[
p(X = 4 | C = green) = \frac{1}{3} \left( \frac{1}{2^\kappa} + \frac{1}{7^\kappa} + \frac{1}{8^\kappa} \right)
\]
and
\[
p(X = 4 | C = red) = \frac{1}{4} \left( \frac{1}{4^\kappa} + \frac{1}{3^\kappa} + \frac{1}{4^\kappa} + \frac{1}{5^\kappa} \right)
\]
Therefore 
\begin{equation}
c^* = \argmax_{c} \left( \frac{3}{7} \; p(X = 4 | C = green),  \quad\frac{4}{7} \; p(X = 4 | C = red) \right)
\end{equation}

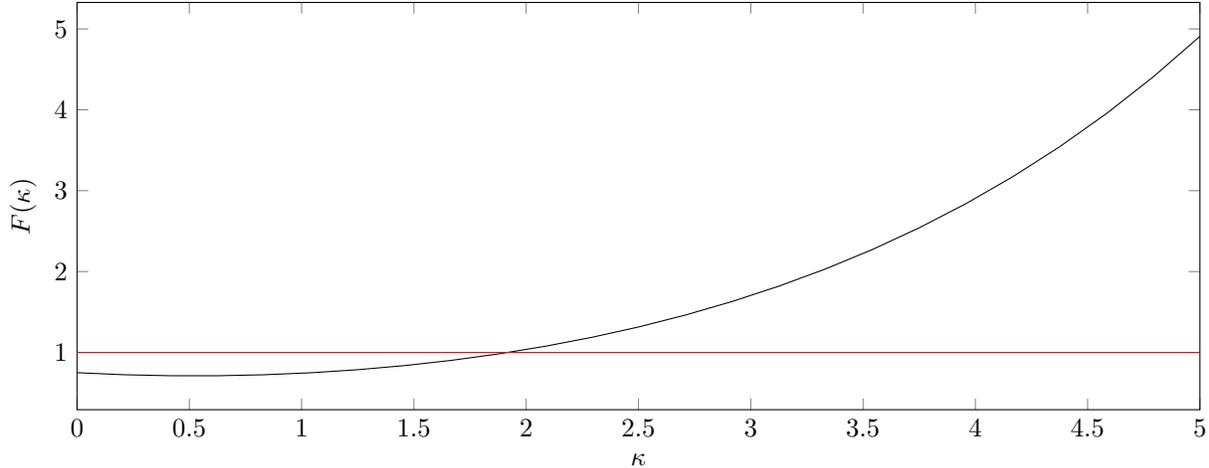
\begin{figure}
\begin{tikzpicture}
\begin{axis}[xlabel = $\kappa$, ylabel = $F(\kappa)$, width=\textwidth, height=7cm, xmin=0, xmax=5]
\addplot[domain=0:5]{(1/(2^x) + 1/(7^x) + 1/8^x)/(1/(4^x) + 1/(3^x) + 1/4^x + 1/5^x)};
\addplot[domain=0:5, color=red]{1};
\end{axis}
\end{tikzpicture}
\caption{Class Probability Ratio versus $\kappa$}
\label{kap}
\end{figure}

In Figure \ref{kap} we plot the ratio of the value for green and the value for red, denoted by $F(\kappa)$, as a function of $\kappa$. If this ratio is greater than one then green should be chosen else red is chosen.

Therefore as long as $\kappa$ is chosen to be sufficiently large (i.e, only consider samples close to the test sample) then the green class is correctly chosen. Note that this holds true for a large range of $\kappa$ values. We will evaluate the performance on all data sets using a single value of $\kappa$ (hence no tuning) and we will also derive the $\kappa$ values that provides the maximum accuracy.

\section{Numerical Results}

In this section we describe the datasets that were used and apply the various classifiers in order to compare performance. We created a GitHub repository, \cite{code}, with the code used for this evaluation so that the reader can replicate and verify the results obtained.

\subsection{Dataset Description}

We use a wide variety of datasets to illustrate the robustness of the proposed approach. These datasets are all available in the University of California at Irvine (UCI) Machine Learning Repository \cite{Dua:2019} and the reader can find additional information on that site. We removed samples with missing attribute values and we encoded categorical values with integer values. No other pre-processing was performed since we wanted to ensure that our results could easily be replicated. A summary of the dataset used is provided in Table \ref{data}.

\begin{table}[!t]
    \centering
    \setlength{\tabcolsep}{4pt}
    \renewcommand{\arraystretch}{1.5}
    \caption{Summary of Dataset Statistics}
    \label{data}
    \begin{tabular}{|l|c|c|c|r|} \hline
    {\bf Name} & {\bf No. of Samples} & {\bf No. of Attributes} & {\bf No. of Classes} & {\bf Citation} \\ \hline
    
    Iris & 150 & 4 & 3 & \cite{fisher1936use}\\ \hline
    Breast Tissue & 106 & 10 & 6 & \cite{estrela2000classification} \\ \hline
    Algerian Forest Fires & 244 & 12 & 2 & \cite{abid2019predicting} \\ \hline
    Credit Approval & 653 & 16 & 2 & \cite{quinlan1987simplifying} \\ \hline
    Wine & 178 & 13 & 3 & \cite{aeberhard1994comparative} \\ \hline
    Breast cancer & 286 & 9 & 2 & \cite{breastcancer}  \\ \hline
    Wine-quality-red & 1599 & 11 & 6 & \cite{wine-red-white} \\ \hline
    Tic-tac-toe & 958 & 9 & 2 & \cite{tictactoe} \\ \hline
    Australian Credit Approval & 690 & 14 & 2 & \cite{quinlan1987simplifying} \\ \hline
    Yeast & 1484 & 9 & 10 & \cite{yeast} \\ \hline
    Raisin & 900 & 7 & 2 & \cite{ccinar2020classification} \\ \hline
    Glass & 214 & 9 & 6 & \cite{glass} \\ \hline   
    Leaf & 340 & 15 & 30 & \cite{leaf} \\ \hline
    Wine-quality-white. & 4898 & 11 & 7 & \cite{wine-red-white} \\ \hline
    Banknote authentication. & 1372 & 4 & 2 & \cite{banknote} \\ \hline   
    Dry Bean & 13611 & 17 & 7 & \cite{KOKLU2020105507} \\ \hline
    Abalone & 4177 & 8 & 3 & \cite{nash1994population}\\ \hline    
    \end{tabular} 
\end{table}

\subsection{Feature Selection}

Note that there are many ways to perform feature selection (\cite{banerjee_2020}) and the only way that one can guarantee the optimal subset of features (for a given training set) is by exhaustive search which is typically computationally expensive. The objective of this paper is to compare our proposed classifier with standard classifiers. We therefore try our best (e.g. using Importance Values, \cite{scikit-learn}) to select the best features for the Gaussian Naive Bayes classifier and then use these selected features for all of the other classifiers. This means, in particular, that there may be another subset of features that are better suited (i.e., higher accuracy) for our approach but we do not use those features in our comparison. The features selected for each dataset are provided in \ref{feature}. The reader can use these to verify our results or, if they find a better subset, can compare our approach using their choice. 

\begin{table}[!t]
    \centering
    \setlength{\tabcolsep}{1pt}
    \renewcommand{\arraystretch}{1.5}
    \caption{Features selected to Maximize accuracy for the Gaussian Naive Bayes Classifier}
    \label{feature}
    \begin{tabular}{|l|c|c|r|} \hline
    {\bf Dataset} & \parbox{1.3in}{\center \bf Accuracy\\(all attributes)} & \parbox{1.3in}{\center \bf Accuracy\\(selected attributes)}& {\bf Selected Attributes}\\ \hline
    
    Iris & 0.953 & 0.959 & [3, 2] \\ \hline
    BreastTissue     & 0.657 & 0.658 & [0, 8, 6, 3, 7, 4,5,1] \\ \hline
    Algerian Forest Fires & 0.944 & 0.956 & [8, 5, 10, 6] \\ \hline
    Credit Approval & 0.787 & 0.787 & [10, 8, 14, 7, 9, 3, 4, 12, 5, 0, 11, 2, 13,1] \\ \hline
    Wine & 0.974 & 0.983 & [0, 2, 3, 6, 9, 10,11, 12] \\ \hline
    Breast-cancer & 0.739 & 0.753 & [3, 4, 5, 8, 2, 6] \\ \hline 
    Wine-quality-red & 0.550 & 0.578 & [10, 1, 6, 9, 4, 0, 7] \\ \hline
    Tic-tac-toe & 0.715 & 0.715 & [0, 1, 2, 3, 4, 5, 6, 7, 8] \\ \hline
    Australian Credit Approval & 0.788 & 0.801 & [9, 7, 13, 8, 6, 4, 5, 3, 11, 2, 12] \\ \hline
    Yeast & 0.512 & 0.548 & [3, 1, 2, 8, 4, 6, 0], \\ \hline
    Raisin & 0.823 & 0.823 & [1, 0, 4, 6, 2, 5] \\ \hline
    Glass & 0.457 & 0.499 & [7, 5, 3, 6] \\ \hline    
    Leaf & 0.717 & 0.734 & [1, 2, 3, 4, 5, 6, 7, 8, 11, 12, 13, 14] \\ \hline
    Wine-quality-white & 0.445 & 0.481 &  [10, 1, 4, 6, 2, 9, 8, 0] \\ \hline
    Banknote-authentication  & 0.839 & 0.872 & [0, 1] \\ \hline
    DryBean & 0.764 & 0.764 & all attributes \\ \hline
    Abalone & 0.519 & 0.533 & [5, 3, 4, 6, 7, 2] \\ \hline
    \end{tabular}
\end{table}

\subsection{Performance Results}

Although we chose optimal features based on the Gaussian Naive Bayes classifier, we will use those features across all classifiers that we ran. In this way the results for the other classifiers can potentially be improved if feature selection was performed specifically for them. We used 10-Fold cross-validation for each dataset with a re-sample size of 10. For the DryBean dataset we used 10-Fold cross-validation but we did not employ re-sampling as the sample size was sufficiently large. We include accuracy values for Laplace Naive Bayes, Gaussian Naive Bayes, $k$-NN with the optimal $k$, the proposed approach for a single value of $\kappa=60$ and finally the proposed approach with the optimal value of $\kappa$. The resulting performance results are given in Table \ref{performance}. Note that the proposed approach with optimal $\kappa$ outperformed all other classifiers. Note that, in the case of Bank-authentication, the optimal features for Gaussian Naive Bayes was (0,1). In this case $k$-NN outperformed the proposed approach. We then decided to find the optimal features for the $k$-NN model for this dataset and this turned out to be all features. However in the case of all features our approach  has the same performance as $k$-NN. 

\begin{table}[!t]
    \centering
    \setlength{\tabcolsep}{5pt}
    \renewcommand{\arraystretch}{1.5}
    \caption {Accuracy (percentage of samples correctly predicted) for the various Models}
    \label{performance}
    \begin{tabular}{|l|c|c|c|c|c|} \hline
    \parbox{1in}{\bf Dataset} & \parbox{0.8in}{\center \bf Laplace\\Naive Bayes} & \parbox{0.8in}{\bf \center Gaussian \\ Naive Bayes} & \parbox{0.8in}{\center \bf{$K$-NN}\\(Optimal $K$)} & \parbox{0.8in}{\center \bf{Proposed}\\($\kappa=60$)} & \parbox{0.8in}{\center \bf{Proposed}\\(Optimal $\kappa$)}\\ \hline
    Iris  & 0.946 & 0.959 & 0.952 (28) & 0.964 & 0.964 (29) \\ \hline
Breast Tissue  & 0.251 & 0.658 & 0.576 (4) & 0.684 & 0.685 (68) \\ \hline
Algerian Forest Fires   & 0.92 & 0.956 & 0.952 (8) & 0.970 & 0.974 (19) \\ \hline
Credit Approval  & 0.854 & 0.788 & 0.682 (32) & 0.831 & 0.869 (5) \\ \hline
Wine   & 0.766 & 0.983 & 0.759 (1) & 0.98 & 0.988 (5) \\ \hline
Breast-cancer  & 0.754 & 0.753 & 0.729 (6) & 0.73 & 0.758 (9) \\ \hline
Wine-quality-red   & 0.597 & 0.578 & 0.601 (1) & 0.671 & 0.686 (36) \\ \hline
tic-tac-toe  & 0.697 & 0.715 & 0.834 (6) & 0.843 & 0.843 (35) \\ \hline
Australian  & 0.847 & 0.801 & 0.714 (10) & 0.826 & 0.871 (5) \\ \hline
Yeast  & 0.502 & 0.548 & 0.579 (1) & 0.565 & 0.591 (32) \\ \hline
Raisin  & 0.543 & 0.823 & 0.843 (7) & 0.861 & 0.861 (60) \\ \hline
Glass  & 0.535 & 0.499 & 0.628 (7) & 0.686 & 0.686 (60) \\ \hline
Leaf   & 0.065 & 0.734 & 0.586 (2) & 0.728 & 0.737 (37) \\ \hline
Wine-quality-white  & 0.543 & 0.481 & 0.602 (1) & 0.678 & 0.692 (42) \\ \hline
Banknote-authentication [0,1] & 0.561 & 0.872 & 0.949 (3) & 0.933 & 0.942 (89) \\ \hline
Banknote-authentication [All] & 0.497 & 0.839 & 0.999 (2) & 0.999 & 0.999 (35) \\ \hline
DryBean   & 0.228 & 0.764 & 0.748 (1) & 0.922 & 0.922 (61)  \\ \hline
Abalone  & 0.507 & 0.533 & 0.537 (45) & 0.546 & 0.545 (27) \\ \hline

Average Score & 0.595 & 0.732 & 0.722  & 0.789 & 0.801  \\ \hline


    \end{tabular}
\end{table}

\subsection{Complexity and Run-Time Analyses}

Although the proposed approach performs well when compared with Naive Bayes classifiers, it does require more computation. Assuming $m$ samples, $n$ attributes and $c$ classes the computational complexity for training of Naive Bayes classifiers is $O(mn)$ while for $k$-NN and our approach it is $O(1)$. However, for predicting a sample the computational complexity is $O(cn)$ for the Naive Bayes classifiers and $O(mn)$ for  $k$-NN  and our approach and so predicting is significantly higher. 

We also did some run time testing for the datasets in \ref{data} for the various classifiers. Our run time testing is implemented using the built-in Python {\tt time} module, specifically, {\tt time.perf\_counter()}. According to the Python documentation the {\tt time.perf\_counter} returns the fractional seconds of a clock with the highest available resolution for measuring short durations.
We found that, on average, it took 2.4 times longer to run the proposed approach (for a single value of $\kappa$) when compered to the time taken for the Laplace Naive Bayes approach. Note that we believe that further computation optimizations can be performed for the proposed approach. We also plan to determine ways to quickly determine good values for $\kappa$. 


\subsection{Parameter Optimization}

\begin{figure}[!t]
\begin{tikzpicture}
\begin{axis}[
    width=\textwidth,
    height=10cm,
    xlabel={$\kappa$},
    ylabel={Accuracy},
    xmin=0, xmax=95,
    ymin=0.5, ymax=1,
    legend style={at={(1,0)},anchor=south east}
]
\addplot[
    color=blue,
    mark=*,
    ]
    coordinates {
(0,0.5628666666666667)
(5,0.9535166666666675)
(10,0.9600000000000009)
(15,0.9689833333333345)
(20,0.9738500000000011)
(25,0.9754833333333345)
(30,0.9758666666666678)
(35,0.9754666666666678)
(40,0.9742333333333345)
(45,0.9726333333333347)
(50,0.9718000000000013)
(55,0.9701833333333347)
(60,0.9701833333333347)
(65,0.9669000000000016)
(70,0.9661000000000017)
(75,0.9640666666666685)
(80,0.9636666666666684)
(85,0.9628500000000018)
(90,0.9620166666666684)
(95,0.9616000000000018)
    };

\addplot[
    color=red,
    mark=*,
    ]
    coordinates {
(0,0.33346320346320335)
(5,0.5436796536796537)
(10,0.6193290043290045)
(15,0.6422077922077923)
(20,0.6499999999999999)
(25,0.6570562770562771)
(30,0.6734632034632033)
(35,0.6833766233766235)
(40,0.6866666666666668)
(45,0.683917748917749)
(50,0.6801948051948054)
(55,0.6820779220779221)
(60,0.685844155844156)
(65,0.6872510822510823)
(70,0.6881818181818181)
(75,0.6904545454545455)
(80,0.689047619047619)
(85,0.6876406926406925)
(90,0.6811471861471862)
(95,0.6792857142857143)
    };

\addplot[
    color=green,
    mark=*,
    ]
    coordinates {  
(0,0.5550724637681159)
(5,0.8710144927536234)
(10,0.8695652173913042)
(15,0.8627536231884054)
(20,0.8531884057971014)
(25,0.8475362318840579)
(30,0.8449275362318839)
(35,0.8405797101449275)
(40,0.8366666666666666)
(45,0.8313043478260869)
(50,0.828695652173913)
(55,0.8279710144927535)
(60,0.8263768115942028)
(65,0.824782608695652)
(70,0.824927536231884)
(75,0.8231884057971013)
(80,0.8213043478260867)
(85,0.8204347826086955)
(90,0.8204347826086957)
(95,0.8192753623188406)
    };
 
 \addplot[
    color=purple,
    mark=*,
    ]
    coordinates {     
(0,0.22533333333333352)
(5,0.9573333333333335)
(10,0.9593333333333335)
(15,0.9600000000000001)
(20,0.9593333333333334)
(25,0.9606666666666668)
(30,0.9640000000000001)
(35,0.9646666666666668)
(40,0.9640000000000001)
(45,0.9640000000000001)
(50,0.9640000000000001)
(55,0.9633333333333335)
(60,0.9640000000000002)
(65,0.962666666666667)
(70,0.9640000000000003)
(75,0.9640000000000003)
(80,0.9626666666666669)
(85,0.9626666666666669)
(90,0.9626666666666669)
(95,0.9626666666666669)
    };

 \addplot[
    color=orange,
    mark=*,
    ]
    coordinates {     
(0,0.3984640522875816)
(5,0.9876143790849681)
(10,0.987189542483661)
(15,0.9787581699346412)
(20,0.9770588235294122)
(25,0.9765032679738567)
(30,0.9781699346405236)
(35,0.9764705882352948)
(40,0.9776470588235301)
(45,0.9787581699346413)
(50,0.9798692810457523)
(55,0.9793137254901968)
(60,0.9804248366013079)
(65,0.9804248366013079)
(70,0.9804248366013079)
(75,0.9804248366013079)
(80,0.9804248366013079)
(85,0.9804248366013079)
(90,0.9804248366013079)
(95,0.9804248366013079)
    };

\legend{Algerian, glass, Australian, Iris, Wine}  
\end{axis}
\end{tikzpicture}
\caption{Accuracy as a function of $\kappa$}
\label{sens2}
\end{figure}
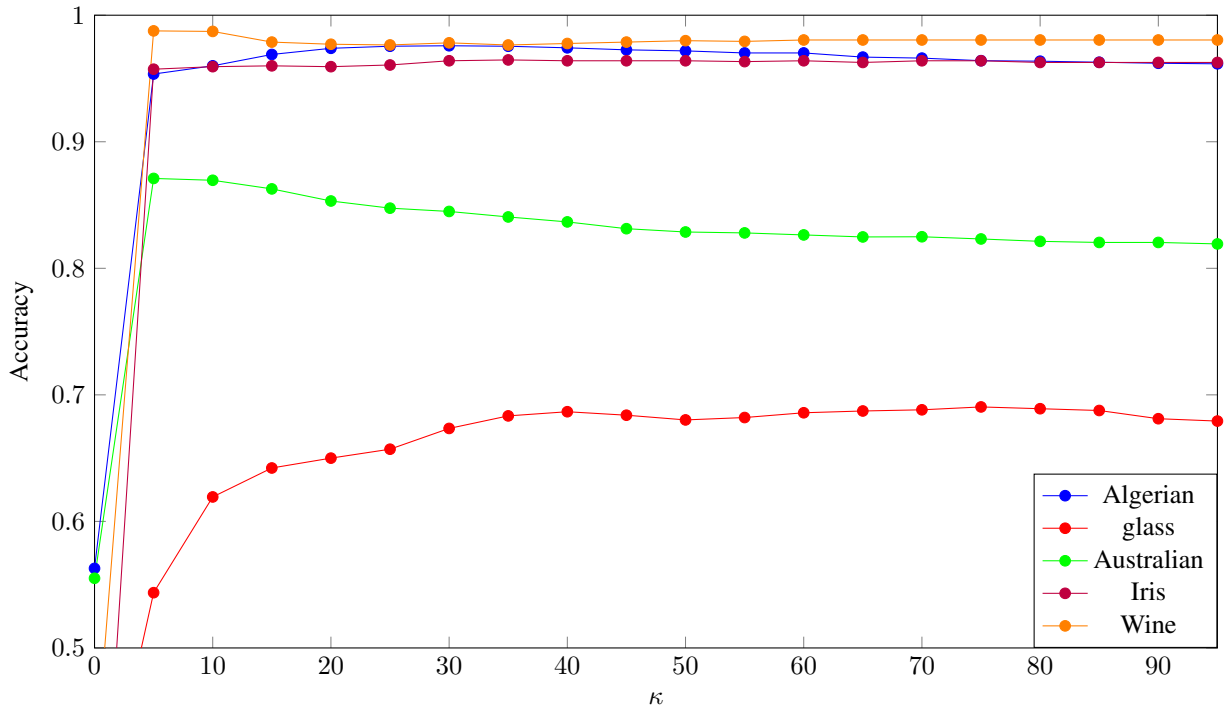

Note that the approach requires a single parameter $\kappa$. However we found, in Table \ref{data},  that even if we used a single value for $\kappa$ (in this case 60), for each dataset we get acceptable results. In fact the average accuracy over all dataset when using $\kappa=60$ is 0.789 while the accuracy when using the optimal $\kappa$ was 0.801. In addition,  without optimizing $\kappa$ our algorithm is successful in 59\% of the datasets. For this exercise we did a linear search over $\kappa$ but we have noticed that in most cases there is a single maximum. In Figure \ref{sens2} we plot the accuracy of the proposed classifier for a subset of datasets as $\kappa$ is varied. We find that, as long as a sufficiently high value of $\kappa$ is used, the result is close to optimal and in fact the approach is robust. 
 
 Note that the case $\kappa=0$ corresponds to averaging over all samples and hence does not take into account the attributes of a test sample. If such attributes are correlated with the class then one would expect this extreme to be non-optimal. The case in which $\kappa$ is very large corresponds to only taking into account samples very close to the test sample. Here the attribute values of the test sample are taken into account but insufficient samples (or none) may exist close to the test sample and so the estimate is not robust leading again to large errors. Therefore we expect $\kappa$ to lie in-between these extremes. We conjecture that the accuracy, as a function of $\kappa$, is approximately concave and we plan to investigate this more in the future when investigating ways of determining good initial guesses for $\kappa$. Not all of the plots are concave but this may be due to the stochastic nature of the problem. Also note that, as the sample size increases, the optimal value of $\kappa$ will also increase since more samples will lie within the neighbourhood of the test sample. Anyway there seems to be a single maximum. If this is indeed the case then the search over $\kappa$ can be significantly reduced.

\section{Discussion}

Across 17 UCI datasets we compared the Laplacian Naive-Bayes, Gaussian Naive-Bayes and $K$-Nearest Neighbors classification algorithms to our proposed method. The  datasets vary in data type and consists of a combination of ordinal, categorical and continuous data. The classification space varies from binary to multiple classes and the size of the feature space also varies significantly. We note in Table \ref{performance} that the average accuracy of the proposed method is 9.43\% higher than the Gaussian approach, 34.6\% higher than the Laplacian approach and 10.1\% higher than $K$-Nearest Neighbors.  These results suggest that the proposed method performs well and is robust when compared with traditional methods. The proposed approach had a higher accuracy in all datasets except Banknote-authentication \cite{banknote}. Features of [0,1] resulted in $K$-NN having an accuracy of 0.949 and our model, 0.942. In this case we tried using all features and tied with $K$-Nearest Neighbours with an accuracy of 0.999. The optimal $\kappa$ was found through a grid search but an arbitrarily high value of $\kappa$ (e.g., 60) gives an average accuracy higher than the other three methods and outperforms them in 58.8\% of the cases. 


%

\section{Conclusions and Future Work}

We present a robust approximation to Bayesian classification using an activation function in the form $(1+x)^{-\kappa}$, where $x$ is a distance metric computed using the euclidean distance between test and training samples. When compared to the Naive-Bayes algorithm (Gaussian and Laplacian) the proposed method surpasses performance or is on par. The proposed method also outperforms the $k$-NN classifier as demonstrated through experimentation. While the results presented in this paper show promise, there is significant room for improvement of the algorithm. Future work can include heuristic methods for determining $\kappa$ or working on reducing run-time for the proposed approach by using a subset of training samples when making a prediction. As the proposed method relaxes the independence assumption of the Naive-Bayes classifier it may be possible to use it for image classification. 

\bibliographystyle{unsrtnat}
\bibliography{nb}

\end{document}